\documentclass[letterpaper, 10 pt, conference]{ieeeconf}  

\IEEEoverridecommandlockouts                              

\usepackage{cite}
\usepackage{graphicx}
\usepackage{booktabs}
\usepackage{multirow}

\title{\LARGE \bf
LGC-Net: A Lightweight Gyroscope Calibration Network for Efficient Attitude Estimation
}

\author{Yaohua Liu$^{1,2,3}$, Wei Liang$^{3}$ and Jinqiang Cui$^{2,*}$ 
\thanks{*corresponding author.}
\thanks{$^{1}$School of Nano-Tech and Nano-Bionics, University of Science and Technology of China, 96 Jinzhai Road, Hefei City, Anhui Province, China, 230026.}%
\thanks{$^{2}$Peng Cheng Laboratory, No.2, Xingke 1st Street, Nanshan, Shenzhen, China, 518055, {\tt\small cuijq@pcl.ac.cn}}%
\thanks{$^{3}$Institute of Nano-Tech and Nano-Bionics, Chinese Academy of Sciences, 398 Ruoshui Road, Suzhou Industrial Park, Suzhou City, Jiangsu Province, China, 215123.}%
}

\begin{document}

\maketitle
\thispagestyle{empty}
\pagestyle{empty}

\begin{abstract}
This paper presents a lightweight, efficient calibration neural network model for denoising low-cost microelectromechanical system (MEMS) gyroscope and estimating the attitude of a robot in real-time. The key idea is extracting local and global features from the time window of inertial measurement units (IMU) measurements to regress the output compensation components for the gyroscope dynamically. Following a carefully deduced mathematical calibration model, LGC-Net leverages the depthwise separable convolution to capture the sectional features and reduce the network model parameters. The Large kernel attention is designed to learn the long-range dependencies and feature representation better. The proposed algorithm is evaluated in the EuRoC and TUM-VI datasets and achieves state-of-the-art on the (unseen) test sequences with a more lightweight model structure. The estimated orientation with our LGC-Net is comparable with the top-ranked visual-inertial odometry systems, although it does not adopt vision sensors. We make our method open-source at: {\tt\small https://github.com/huazai665/LGC-Net}

\end{abstract}

\section{INTRODUCTION}
Low-cost inertial measurement units (IMUs) based on microelectromechanical system (MEMS) have been popularly used in robots navigation due to their small size and low power consumption. IMUs can provide the attitude and position information by integrating 3-axis gyroscope and acceleration measurements, i.e., angular velocity and linear acceleration, without external equipment or signals. However, the low-cost MEMS IMUs often suffer from complex, nonlinear, time-varying noise and random errors\cite{rehder2016extending}, \cite{rohac2015calibration}. Therefore, a more efficient calibration algorithm will be crucial in a more robust and accurate attitude estimation.  

Discrete calibration and system-level calibration are two classical methods used in inertial navigation to correct IMU errors\cite{huang2022mems}. Most discrete calibration methods model the IMU measurement output as a linear polynomial equation for the systematic error, such as constant bias, scale factor, and axes misalignment error. The corresponding coefficient parameters are calculated relying heavily on high-precision external equipment, i.e., 3-axis turntable\cite{tedaldi2014robust, ghanipoor2020toward, furgale2013unified, lu2017all}. However, the high-accuracy turntables are usually expensive and heavy, and the discrete IMU output model ignores the high-order coupling term and time-varying characteristics. The system-level calibration can correct IMU error by reducing inertial navigation result error based on the Kalman filter. Although the system-level calibration method reduces the dependence on external equipment and is suitable for online calibration, it is hard to denoise the nonlinear and non-Gaussian random error.

\begin{figure}
    \centering
    \includegraphics[width=.99\linewidth]{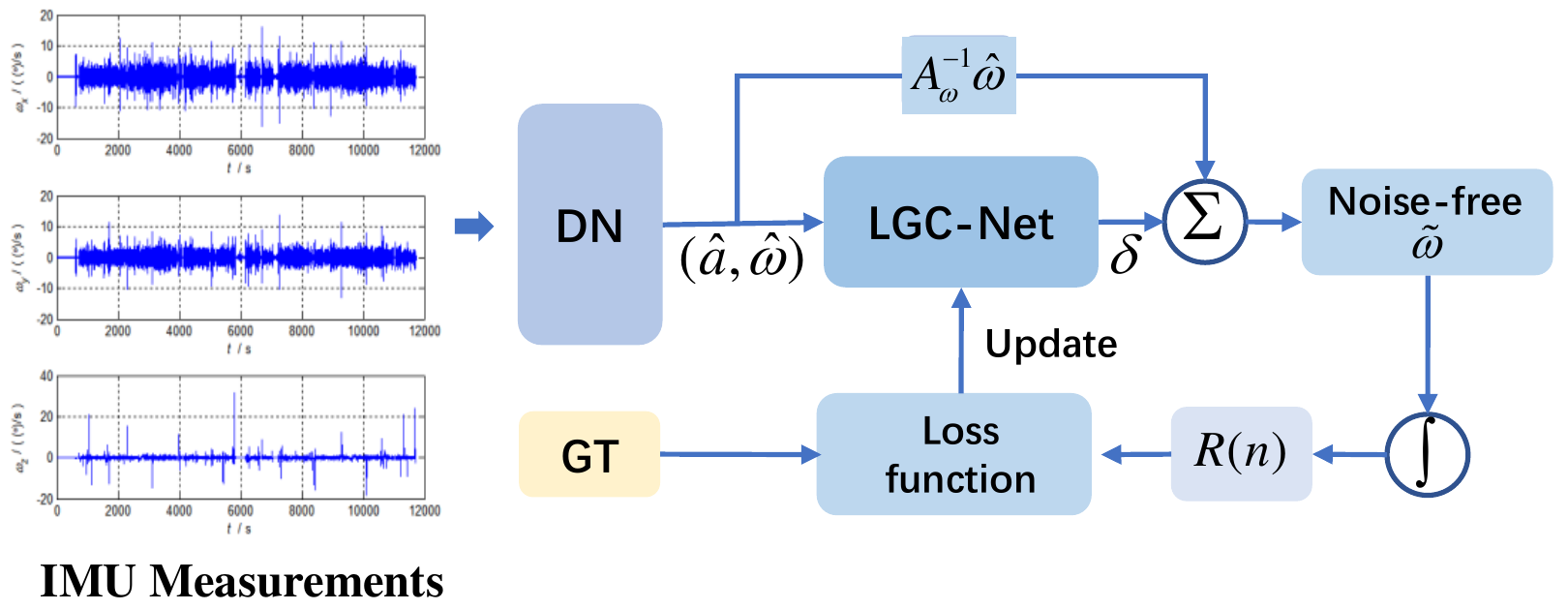}
    \caption{Framework of the proposed IMU calibration}
    \label{fig:framework}
\end{figure}
Recently, with the development of deep learning technologies, learning-based methods have been introduced into the visual-inertial navigation field to improve navigation and positioning accuracy \cite{chen2018ionet,herath2020ronin,esfahani2019orinet,rao2022ctin,herath2022neural}. In addition, the use of deep learning technology to calibrate IMU error has begun to attract researchers' attention and the existing research shows that the MEMS IMU calibration based on deep learning is efficient \cite{brossard2020denoising,huang2022mems,engelsman2022data}. However, they do not consider sectional features and long-range dependencies simultaneously, and the models are not lightweight enough to be deployed to edge devices.  

To address these problems, we propose a lightweight and efficient deep calibration network for low-cost MEMS gyroscope and open-loop orientation estimation only using the calibrated gyroscope measurements in real-time. The contributions of the paper are summarized as follows:
\begin{enumerate}
    \item A lightweight calibration network called LGC-Net based on depthwise separable convolution and large kernel attention is proposed that takes raw MEMS IMU data as the input and estimates the calibration compensation components.
    \item The developed LGC-Net can achieve state-of-the-art attitude estimate performance only using calibrated gyroscope data with fewer model parameters than the existing learning-based methods. 
    \item A series of experiments on the EuRoC and TUM-VI public datasets are performed to verify the performance of the proposed method. To promote further research,  we make our algorithm open-source.
\end{enumerate}

\section{RELATED WORK}
MEMS IMU calibration can be defined as a process that compares instrument outputs with known reference information to determine coefficients that force the output to agree with the reference information across the desired range of output values. All systematic errors can be compensated when only the significant errors are considered, and some random errors can be estimated by the navigation algorithm and compensated online\cite{noureldin2013fundamentals}. U. Qureshi et al.\cite{qureshi2017algorithm} proposed an in-field calibration of a MEMS IMU, which requires no external equipment and utilizes gravity signal as a stable reference.
Aiming at the swaying base environment, Dai et al.\cite{dai2018full} improved the filtering method for 3-axis rotational inertial navigation system by taking the velocity errors as the reference. 
With the consideration of external perturbations, Peng et al.\cite{peng2021design} developed a low-cost, lightweight, and portable IMU calibration embedded platform and proposed an iterative weighted Levenberg-Marquardt algorithm to deal with perturbations. While the errors of scale factors, misalignment, and offsets constantly vary with time and environment, these calibration methods usually consider the parameters as constant values.

In order to calibrate and reduce the random error, researchers have proposed many representative denoising techniques for MEMS IMUs. The denoising methods are mainly based on statistical principles, including autoregressive moving average method (ARMA)\cite{song2018improved}, Allan variance\cite{zhang2018application}, Kalman filter\cite{zhang2016dual}, and wavelet transformation\cite{yuan2015improved}. The ARMA method is mainly used to analyze and study a group of random data arranged in sequence and establish mathematical models of various orders according to different error sequences. However, this method cannot identify random errors one by one, making it difficult to distinguish the error sources. To solve the above problem, Allan variance is proposed to identify various random errors and separate them into five parts: quantization noise, angle random walk, bias instability, rate random walk, and rate ramp. Kalman filter is an efficient linear quadratic estimator which can estimate gyroscope output angular velocity via a series of observed measurements with noise. However, since the actual MEMS IMU error is usually too complex to build an accurate mathematics model, the Kalman filter has poor performance for denoising random errors. Among the statistical methods, the wavelet transform method is a standard method for reducing the high-frequency part of the gyroscope error but it cannot remove the low-frequency errors.

For the visual-inertial navigation system, Furgale et al.\cite{furgale2013unified} presented a new calibration framework for jointly estimating the temporal offset and spatial displacement between camera and IMU and made it open-source, called Kalibr toolbox. Rehder et al.\cite{rehder2016extending} further improved the Kalibr toolbox for spatially calibrating multiple IMUs following an inertial noise model. Jung et al.\cite{jung2020observability} proposed a self-calibrated visual-inertial odometer(VIO) to estimate IMU scale factor and misalignment using an extended Kalman filter-based pose estimator. As to the problem of the necessity of online IMU intrinsic calibration, Yang et al.\cite{yang2020online} performed observability analysis for the visual-inertial navigation system and first identified six primitive degenerate motion profiles for IMU intrinsic calibration. However, these methods rely heavily on the IMU error model, resulting in poor correction accuracy in some challenges.

In recent years, deep learning has been introduced into the inertial odometer, such as OriNet\cite{esfahani2019orinet}, IONet\cite{chen2018ionet}, TLIO\cite{liu2020tlio}, A2DIO\cite{wang2022a2dio}, CTIN\cite{rao2022ctin}, all of which achieve great positioning performance than traditional methods. However, the use of deep learning technology to calibrate IMU errors has just started, and the published research results are still rare. The key idea is that a deep neural network is trained to mine internal implicit relationships with a lot of raw IMU data. In \cite{jiang2018mems}, a long short-term memory (LSTM) neural network is employed to filter the MEMS gyroscope outputs by treating the signals as time series, showing that the denoising scheme effectively improves MEMS IMU accuracy. To further explore the effect of LSTM in denoising the MEMS IMU, some hybrid deep recurrent neural networks, including LSTM and gated recurrent unit (GRU), are evaluated for MEMS IMU with static and dynamic conditions\cite{han2021hybrid}. Chen et al.\cite{chen2018improving} first present a novel deep learning-based method to simultaneously remove systematic and random errors using a Convolutional Neural Network (CNN). In 2020, Brossard et al.\cite{brossard2020denoising} designed a CNN to estimate attitude with only a commercial IMU, and the test results outperformed the state-of-the-art on the unseen sequences of EuRoC and TUM-VI datasets. Recently, Huang et al.\cite{huang2022mems} adopted a temporal convolutional network to construct the MEMS IMU gyroscope output model and improve the attitude and position accuracy of the inertial navigation solution.

\section{METHOD}
We focus on the low-cost MEMS IMU composed of a 3-axis accelerometer and a 3-axis gyroscope, which can provide the linear acceleration ${\hat a}_n$ and angular velocity ${\hat \omega }_n$ information separately in the body frame. Our goal is to estimate the correction parameters of the IMU using deep neural network and directly calculate the attitude using noise-free angular velocity values. 
\begin{figure}[h]
    \centering
    \includegraphics[width=.95\linewidth]{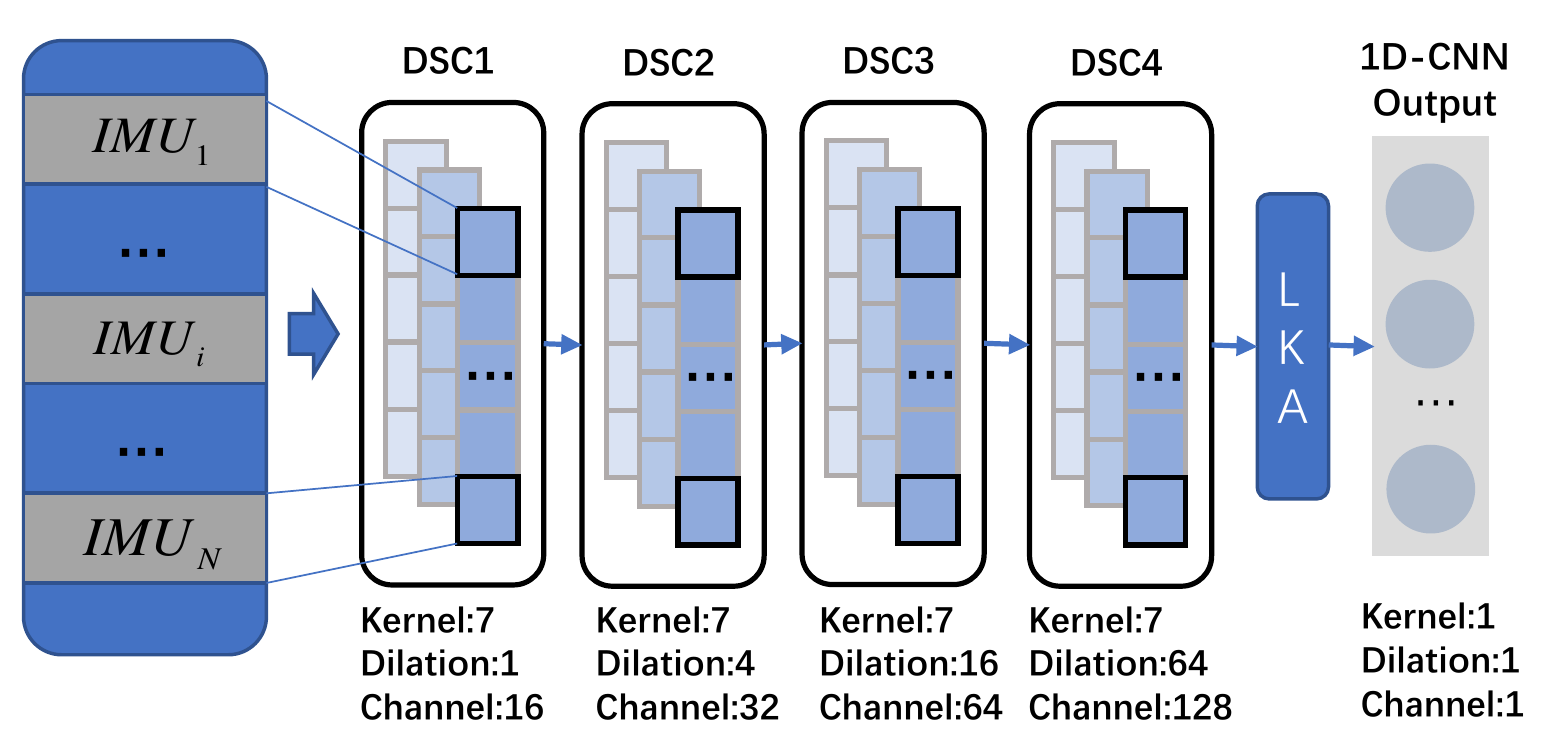}
    \caption{The proposed LGC-Net structure}
    \label{fig:lgc-net}
\end{figure}

\subsection{MEMS IMU Output and Attitude Update Models}
For a typical low-cost MEMS IMU, the inertial measurements can be expressed as follows,
\begin{equation}
    \left[ {\begin{array}{*{20}{c}}
{{{\hat a}_n}}\\
{{{\hat \omega }_n}}
\end{array}} \right] = A\left[ {\begin{array}{*{20}{c}}
{{a_n}}\\
{{\omega _n}}
\end{array}} \right] + \left[ {\begin{array}{*{20}{c}}
{b_n^a}\\
{b_n^\omega }
\end{array}} \right] + \left[ {\begin{array}{*{20}{c}}
{\eta _n^a}\\
{\eta _n^\omega }
\end{array}} \right]
\end{equation}
where $a_n$ and $\omega _n$ are the actual values of the linear acceleration and angular velocity at timestamp $n$; ${b_n^a}$ and ${\eta _n^a}$ are the bias and additive zero-mean Gaussian noises for accelerometer, and ${b_n^\omega }$ and ${\eta _n^\omega }$ are for gyroscope; The intrinsic calibration matrix $A$ can be further represented as \cite{rehder2016extending},
\begin{equation}
   {A = }\left[ {\begin{array}{*{20}{c}}
{{S_a}{M_a}}&{{C_ \times }}\\
{{0_{3 \times 3}}}&{{S_\omega }{M_\omega }}
\end{array}} \right] \approx {I_6}
\end{equation}
where $S_a$ and ${S_\omega }$ represent the scale factors,  ${{M_a}}$ and ${M_\omega }$ are axis misalignments, respectively. Both  $S$ and $M$ are approximately equal to the identity matrix. ${C_ \times }$ is the coefficient matrix representing the effect of linear acceleration on gyroscope measurement, i.e., g-sensitivity\cite{brossard2020denoising}. Thus, the acceleration output data would be considered part of the neural network input to eliminate the g-sensitivity error.

The attitude of a carrier can be calculated by integrating orientation increments from the gyroscope measurement as follows,
\begin{equation}\label{attitude_update}
R(n) = R(n - 1)\exp ({\theta _n})
\end{equation}
where $R(n)$ is the rotation matrix of the carrier frame relative to the inertial frame at timestamp $n$ and $\theta _n$ is the orientation increments with the exponential map $\exp ( \cdot )$ in the $SO(3)$, which can be inferred as,
\begin{equation}
    {\theta _n} = {{\hat \omega }_n}dt
\end{equation}
\begin{equation}
    \exp ({\theta _n}) = I + \frac{{\sin {\theta _n}}}{{{\theta _n}}}[{\theta _n} \times ] + \frac{{1 - \cos {\theta _n}}}{{\theta _n^2}}{[{\theta _n} \times ]^2}
\end{equation}
where $[{\theta _n} \times ]$ is the antisymmetric matrix of $\theta _n$. Thus, the error present in $R(n - 1)$ will be transferred to the $R(n)$ according to Eq.\ref{attitude_update}.

\subsection{LGC-Net for Gyroscope Measurement Calibration}
Through the analysis of the MEMS IMU output model, the gyroscope measurements after error calibration can be represented as,
\begin{equation}\label{cali}
    {{\tilde \omega }_n} = A_\omega ^{ - 1}(\hat \omega_n  - b_n^\omega  - \eta _n^\omega ) = A_\omega ^{ - 1}\hat \omega  + \delta 
\end{equation}
where ${\tilde \omega }_n$ is the gyroscope data after error compensation, $A_\omega$ contains scale factor and axis misalignment errors for gyroscope, and we define $\delta  = A_\omega ^{ - 1}(b_n^\omega  + \eta _n^\omega)$ as the gyroscope correction which is a time-varying error due to the bias is random and time-varying.

As shown in Eq.\ref{cali}, $A_\omega$ and $\delta$ are essential for IMU calibration and attitude estimation. According to their error characteristics, we design the calibration model based on LGC-Net illustrated in Fig.\ref{fig:framework}. Firstly, in order to accelerate the convergence speed of the network model, we conduct data normalization (DN) preprocessing on the IMU data to make it obey normal distribution. The normalized IMU sequences are then split into small windows with $N$ samples, which indicates our model can capture the inhere inertial information from $N$ data at one time. The scale factor and axis misalignment errors in $A_\omega$ are not easily affected by the external environment, thus $A_\omega$ can be regarded as a constant parameters initialized at identity matrix and optimized during training in our model. Since $\delta$ is nonlinear time-varying affected by many factors, we design the LGC-Net to predict $\delta$ using IMU data in a time window $N$, and the network can be expressed as,
\begin{equation}
    {\delta _t} = f(IM{U_{t - N}},...,IM{U_t})
\end{equation}
where $f( \cdot )$ is the nonlinear function learned by LGC-Net, The learned $f( \cdot )$ can capture the temporal multi-scales features and regress the gyroscope corrections.

Afterward, the noise-free measurement can be obtained by summing $A_\omega ^{ - 1}\hat \omega$ and $\delta$, and the attitude can be estimated by open-loop integrating the calibrated gyroscope outputs. Furthermore, the parameters of the LGC-Net are also updated by calculating the loss function between ground truth and estimated attitudes. Thus, the well-trained network can correct the MEMS gyroscope measurements through repeated training and accomplish more accurate attitude estimation.

\subsection{Network Structure}
Normalized IMU measurements can be sent to the LGC-Net for further processing. In order to make the network lightweight, we leverage depthwise separable convolution (DSC) inspired by mobilenet\cite{howard2017mobilenets} and large kernel attention (LKA)\cite{guo2022visual} to infer a gyroscope correction based on a time window of $N$. Considering mining the correlation of IMU data as much as possible and the size of GPU memory, we finally choose to set $N$ to 16000 after many tests. Specially, the LGC-Net is composed of four DSC layers, one LKA block, and one output layer. The detail of the LGC-Net and the well-chosen configuration of each layer is presented in Fig.\ref{fig:lgc-net}. We also set $A_\omega$ as a trainable parameter to keep tuning it. According to the mathematical model in Eq.\ref{cali}, the parameters of LGC-Net can be optimized by the carefully designed loss function.
\begin{figure}[h]
    \centering
    \includegraphics[height=2.2cm, width=7.3cm]{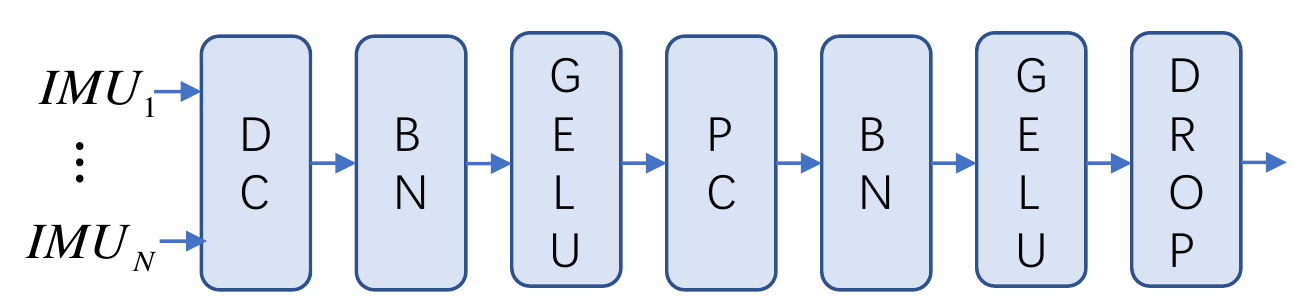}
    \caption{The details of the depthwise separable convolution}
    \label{fig:dsc}
\end{figure}

\begin{figure}[h]
    \centering
    \includegraphics[width=.9\linewidth]{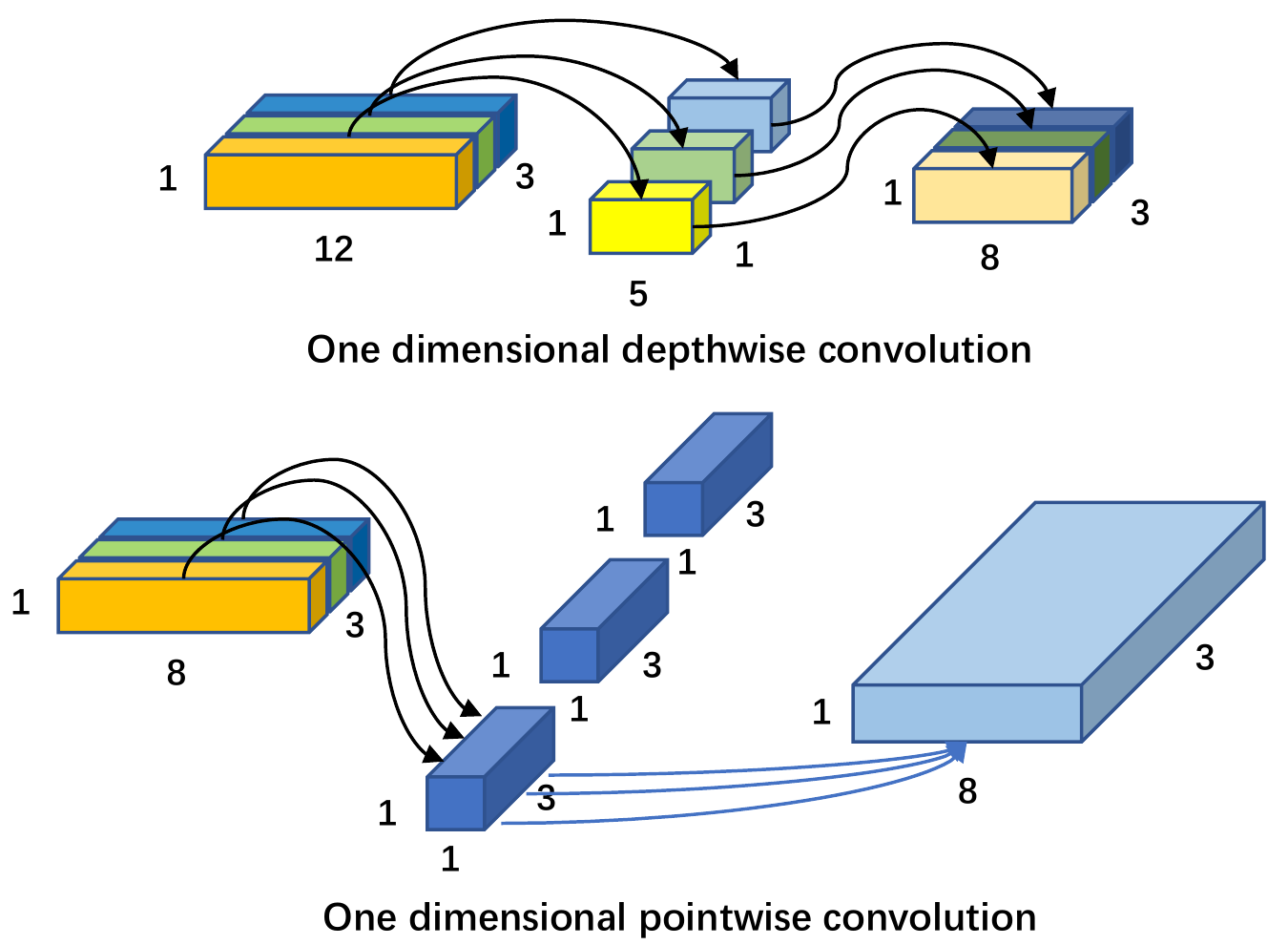}
    \caption{Depth-wise and Point-wise convolution for one dimensional data}
    \label{fig:dpc}
\end{figure}

As shown in Fig.\ref{fig:dsc}, the DSC is mainly composed of depthwise convolution (DC), batch normalization (BN)\cite{ioffe2015batch}, GELU activation function\cite{ramachandran2017searching}, pointwise convolution (PC) and dropout layer. The first step of the designed LGC-Net is to extract the spatial feature map of the MEMS IMU by the DC layer, where the number of feature map channels is the same as the number of input channels. Since DW does not effectively use the feature information of different channels in the same spatial location, we adopted the PC layer to combine the DW's feature map in the depth direction to generate a new feature map. In order to reduce network overfitting and increase nonlinear mapping capability, we design a BN layer and a GELU activation function between DW and PC layers, and a dropout layer is added at the end. The specific DW and PW processes for one-dimensional data are illustrated in Fig.\ref{fig:dpc}. A standard convolution layer can be converted into a DW layer and a PW layer, where one convolution kernel of the DW is responsible for one channel, and one channel is only convoluted by one convolution kernel. The DW can reduce the network parameters but can not increase the number of feature map channels. Thus, the PW is designed as a standard convolution layer with kernel size 1 to increase the depth of the feature map without changing the size. Compared with standard convolution, the DSC can reduce the number of parameters by two-thirds.

\begin{figure}[h]
    \centering
    \includegraphics[width=.3\linewidth]{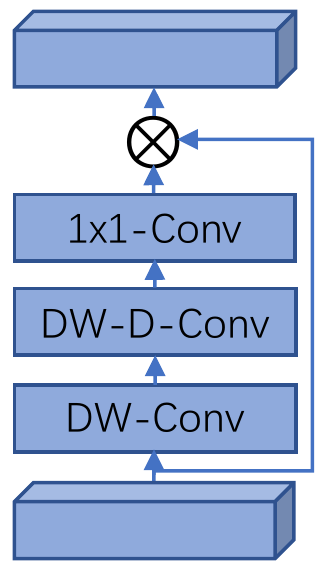}
    \caption{The structure of large kernel attention}
    \label{fig:lka}
\end{figure}

The LKA is a new attention mechanism to enhance the neural network's performance proposed in the visual attention network, which far outperforms the state-of-the-art CNN and visual transformer\cite{dosovitskiy2020image} in terms of image classification, object detection, and semantic segmentation. The LKA can combine the advantages of convolution and self-attention, including local structural information, long-range dependence, and adaptability thanks to the carefully designed structure. The detailed structure of the one-dimensional LKA is shown in Fig.\ref{fig:lka}, which contains a depthwise convolution (DW-Conv), a depthwise dilation convolution (DW-D-Conv), and a 1x1 convolution (1x1 Conv). Since the gyroscope correction values should consider both the previous and instantaneous IMU outputs simultaneously, we embed the LKA behind the DSC4 layer in LGC-Net to capture long-range relationships with fewer computational loads and generate new attention maps according to attention weight. Significantly, the whole LKA can be summarized as,
\begin{equation}
    Att = Con{v_{1 \times 1}}(DW - D - Conv(DW - Conv(F)))
\end{equation}
\begin{equation}
    Output = Att \otimes F
\end{equation}
where $F$ is the input features computed by four DSC layers, attention denotes the attention map, the value of which indicates the importance of each feature. $\otimes$ represents element-wise product. In the end, the $1 \times 1$ convolution with one output channel is selected as the output layer to transform the feature dimension in the hidden layer into the output dimension.

\subsection{Loss Function}
Since the sampling frequency of the IMU is generally 200Hz, calculating directly the difference between the actual angular velocity from the ground truth system and estimated values is unsuitable for real robotic systems. Considering the above reasons, we adopt the incremental calculation method to construct the loss function and design it as the distance between the ground truth increments and integrated increments from the LGC-Net output. The integrated increments can be expressed as follow,
\begin{equation}
    \delta {R_{n,n + t}} = R_n^T{R_{n + t}}
\end{equation}
In order to enhance the robustness for ground truth outliers, we leverage the log hyperbolic cosine (Log-cosh) loss for LGC-Net. Compared with L1 (mean square error) and L2 (mean absolute error) loss, log-cosh loss combines the advantages of the above two losses, decreasing the gradient around the minima. The loss function is defined as,
\begin{equation}
    {L} = \sum\limits_{n = 1}^t {Log - \cosh (\log (\delta {R_{n,n + t}}\delta \tilde R_{n,n + t}^T))} 
\end{equation}
where $\log ( \cdot )$ is the $SO(3)$ logarithm map, and the designed loss approximately equals ${(\log (\delta {R_{n,n + t}}\delta \tilde R_{n,n + t}^T))^2}/2$ for the small loss, and to $\left| {\log (\delta {R_{n,n + t}}\delta \tilde R_{n,n + t}^T)} \right| - \log (2)$ for the large loss.

\begin{table}[h]
\centering
\caption{Training and test sets of EuRoC dataset}\label{EuRoC}
\begin{tabular}{cc}
\hline
\toprule[1pt]
Train dataset     & Test dataset      \\ \hline
MH\_01\_easy      & MH\_02\_easy      \\
MH\_03\_medium    & MH\_04\_difficult \\
MH\_05\_difficult & V1\_01\_easy      \\
V1\_02\_medium    & V2\_02\_medium    \\
V2\_01\_easy      & V1\_03\_difficult \\
V2\_03\_difficult &                   \\ 
\bottomrule[1pt]
\end{tabular}
\end{table}

\begin{table*}[h]
\centering
\caption{Attitude estimation results of different methods on the EuRoC test sequences}\label{EuRoC_result}
\begin{tabular}{ccccclc}
\hline
\toprule[1pt]
EuRoC test sequence & VINS-Mono & Open-VINS     & Raw IMU & OriNet & DIG  & Proposed method \\ \hline
MH\_02\_easy        & 1.34      & \textbf{1.11} & 146     & 5.75   & 1.39 & 1.14            \\
MH\_04\_difficult   & 1.44      & 1.60          & 130     & 8.85   & 1.40 & \textbf{1.32}   \\
V1\_01\_easy        & 0.97      & \textbf{0.80} & 71.3    & 6.36   & 1.13 & 2.81            \\
V2\_02\_medium      & 4.72      & 2.32 & 119     & 14.7   & 2.70 & \textbf{1.98}            \\
V1\_03\_difficult   & 2.58      & \textbf{1.85} & 117     & 11.7   & 3.85 & 3.99           \\
Average             & 2.21      & \textbf{1.55} & 125     & 9.46   & 2.10 & 2.25           \\ 
\bottomrule[1pt]
\end{tabular}
\end{table*}
\section{Experiments}
In order to evaluate the performance of the proposed network model, two different public datasets are tested. Further, we have compared the experiment results with other attitude estimation methods.

\subsection{Dataset Descriptions and Training Details}
The experiment is performed on the EuRoC dataset\cite{burri2016euroc} and the TUM-VI dataset\cite{schubert2018tum}. The EuRoC dataset is a visual-inertial dataset collected from a micro aerial vehicle, which contained 11 flight trajectories of 2-3 minutes in machine hall and vicon room environments. An uncalibrated ADIS16448 IMU is installed in the micro aerial vehicle to sample the inertial data at 200 Hz. In the EuRoC dataset, the ground truth is provided by a laser tracker and Vicon motion capture system, which is accurately time-synchronized with the IMU. We split data into training, validation, and test sets to better tune the neural network parameters. The training and test sequences are summarized in Tab.\ref{EuRoC}, where the first 50 s of the training set is used for training, and the remaining data are used for validation.
\begin{figure}[h]
    \centering
    \includegraphics[width=.95\linewidth]{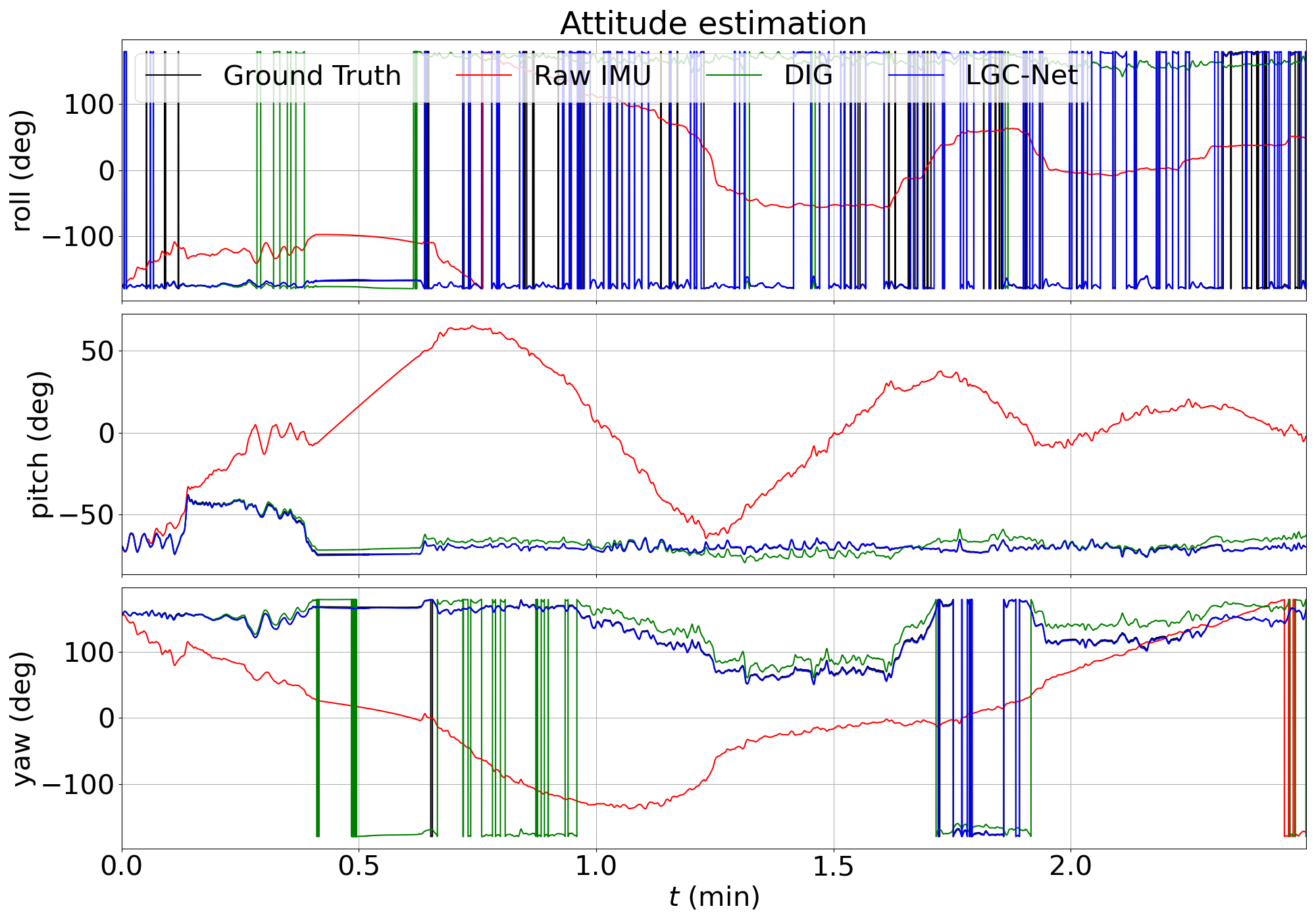}
    \caption{Attitude estimates on the test sequence machine-hall-02}
    \label{fig:MH02}
\end{figure}

The TUM-VI dataset includes inertial and visual data collected from a hand-held device in different scenes. The IMU data is obtained from the cheap BMI160 IMU with a 200 Hz sampling frequency, and the ground truth is generated via the motion capture system. Similarly, we choose the first 50 s of the sequences from $room 1$, $room 3$, and $room 5$ scenes as the training sets, the remaining parts as the validation sets, and the test sets are $room 2$, $room 4$, and $room 6$ sequences.

For the training of our LGC-Net, we use a Dell server computer equipped with an Nvidia GeForce GTX 2080Ti GPU with 11 GB memory.
The framework of LGC-Net is implemented based on PyTorch. The Adam optimizer\cite{kingma2014adam} with the initial learning rate of 0.0008 is used in the training process. The learning rate will be automatically adjusted when the loss changes very slowly. To reduce overfitting, we set the weight decay and dropout parameter are 0.1, respectively, and added a Gaussian noise in the raw IMU measurements to increase the diversity of data based on data augmentation technology during each training epoch. We train the proposed model for 4000 epochs with a short time, i.e., 185 s.

\subsection{Calibration Performance Evaluation on EuRoC Dataset}
To quantitatively evaluate the performance of the LGC-Net, a set of attitude estimation methods are compared. We choose the absolute orientation error (AOE) as the evaluation metric, indicating the similarity between the orientation estimates and the ground truth. Specially, it can be computed by openvins evaluation tool\cite{geneva2020openvins} and expressed as,
\begin{equation}
    AOE = \sqrt {\frac{1}{M}\sum\limits_{n = 1}^M {\log (R_n^T{{\tilde R}_n})} } 
\end{equation}
where $M$ is the sequence length.
\begin{table*}[h]
\centering
\caption{Attitude estimation results of different methods on the TUM-VI test sequences}\label{TUM_result}
\begin{tabular}{cccclc}
\hline
\toprule[1pt]
TUM-VI test sequence & VINS-Mono     & Open-VINS & Raw IMU & DIG  & Proposed method \\ \hline
room 2               & \textbf{0.60} & 2.47      & 118     & 1.31 & 1.60            \\
room 4               & \textbf{0.76} & 0.97      & 74.1    & 1.48 & 1.81            \\
room 6               & \textbf{0.58} & 0.63      & 94.0    & 1.04 & 1.08            \\
Average              & \textbf{0.66} & 1.33      & 95.7    & 1.28 & 1.50            \\ 
\bottomrule[1pt]
\end{tabular}
\end{table*}

The attitude can be calculated by using the corrected gyroscope data and compared with a series of methods, including the following,
\begin{enumerate}
    \item Raw IMU, attitude is estimated using uncalibrated IMU.
    \item OriNet\cite{esfahani2019orinet}, an attitude estimation model based on recurrent neural networks.
    \item Denoising IMU Gyro (DIG)\cite{brossard2020denoising}, open-loop estimating attitude using the calibrated IMU 
    \item VINS-Mono\cite{qin2018vins}, a visual-inertial odometer system with a monocular camera.
    \item Open-VINS\cite{geneva2020openvins}, a filter-based visual-inertial odometer system that achieves a great estimation performance.
\end{enumerate}

The results of attitude estimation with different methods on the test sequence Machine-hall-02 and all test errors are given in Fig.\ref{fig:MH02} and Tab.\ref{EuRoC_result}. As illustrated in Fig.\ref{fig:MH02}, the attitude results from the uncalibrated raw IMU deviate from ground truth in less than 10 s, which cannot be used for robot positioning and navigation. Our proposed method is comparable to the DIG, and both methods can track ground truth well. Compared with the other five methods, our proposed method achieves the best performance on the MH-04-difficult and V2-02-medium test sequence, the AOEs of which are 1.32 deg and 1.98 deg at Tab.\ref{EuRoC_result} separately. The average AOE of our method dropped by 76.2 $\%$ when compared to OriNet. Although our algorithm only using an IMU, the results show that it can compete with VINS-Mono and Open-VINS, which verifies that the proposed model is more reliable and effective for gyroscope error calibration. 

\subsection{Calibration Performance Evaluation on TUM-VI Dataset}
\begin{figure}[h]
    \centering
    \includegraphics[width=.95\linewidth]{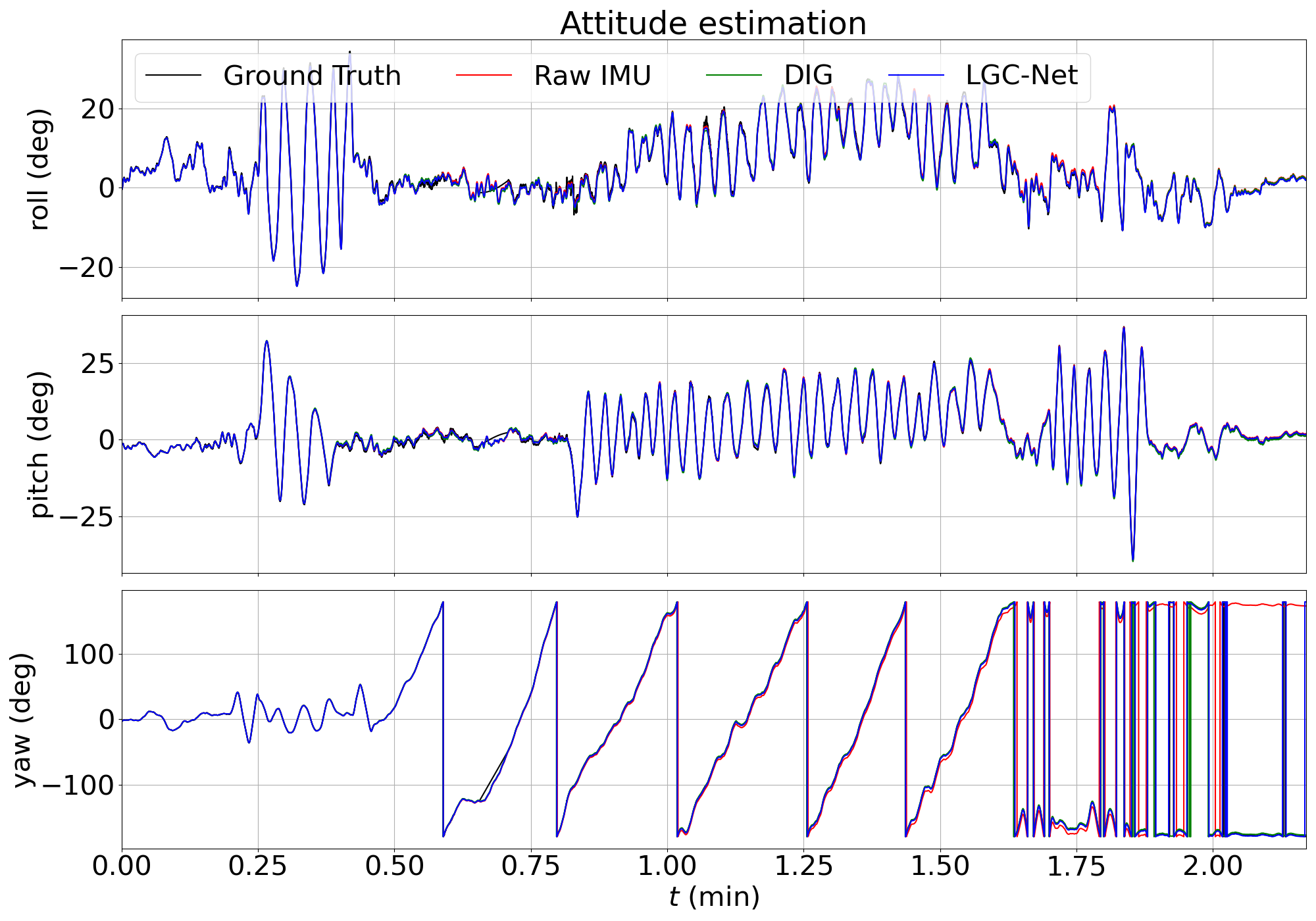}
    \caption{Attitude estimates on the test sequence Room-02}
    \label{fig:room2}
\end{figure}

The attitude estimation is performed in the TUM-VI dataset to validate the proposed method further. Fig.\ref{fig:room2} shows the calculated orientation angles of ground truth, raw IMU, DIG, and LGC-Net. Similar to the EuRoC dataset, the attitude angles from uncorrected IMU have the maximum error, and other method curves are close to the ground truth. 

As shown in Tab.\ref{TUM_result}, the average AOE of raw IMU results is 95.6 deg. The VINS-Mono, based on IMU and camera sensors, has the best performance, the average AOE of which is 0.66 deg. The average AOE of our method is basically at the same level as that of Open-VINS and DIG, i.e., 1.5 deg. Since OriNet does not publish the code and test results on TUM-VI, we cannot compare and analyze them. However, our network model's whole number of trainable parameters is 38146, and the network size is only 0.038 MB. In DIG and \cite{huang2022mems}, the network model parameters are 77052 and 175976, respectively. Compared with them, our model size is reduced by 52.5 $\%$ and 94.6 $\%$. As for OriNet, it uses the recurrent neural network to learn calibration compensation components and needs more computational load consumption. What is more, our model has an extremely small number of parameters compared to visual-inertial learning methods, where IMU processing only requires more than 2600000 parameters\cite{almalioglu2022selfvio}. In the inference phase, our trained network model takes about 1 ms to estimate one test sequence result and has the real-time ability.

\subsection{Ablation Study}
We evaluate the attitude estimation performance of removing the LKA from LGC-Net to demonstrate the effectiveness of the proposed LGC-Net design choice of the LKA. The AOE for the EuRoC and TUM-VI test set of all the ablation experiments are shown in Tab.\ref{ablation}. Applying LKA in LGC-Net has a lower error than without any attention mechanism after the DSC layers, the average AOEs of which are 1.97 deg and 5.73 deg, respectively. The ablation experiments show that all components in the LGC-Net are effective.  

\begin{table}[h]
\centering
\caption{Comparing with LGC-Net and LGC-Net without LKA}\label{ablation}
\begin{tabular}{ccc}
\hline
\toprule[1pt]
Test sequence     & LGC-Net & LGC-Net without LKA \\ \hline
MH\_02\_easy      & 1.14    & 3.87                \\
MH\_04\_difficult & 1.32    & 4.52                \\
V1\_01\_easy      & 2.81    & 7.54                \\
V2\_02\_medium    & 1.98    & 4.83                \\
V1\_03\_difficult & 3.99    & 11.63               \\ \hline
room 2            & 1.60    & 4.05                \\
room 4            & 1.81    & 5.16                \\
room 6            & 1.08    & 4.23                \\ \hline
average           & 1.97    & 5.73                \\
\bottomrule[1pt]
\end{tabular}
\end{table}

\section{Conclusion}
A lightweight gyroscope calibration network for efficient attitude estimation is proposed in this paper. The devised deep neural network achieves outstanding accurate attitude estimates with only a low-cost MEMS IMU, which outperforms state-of-the-art visual-inertial system and Orinet in some test sequences with fewer model parameters. We leverage depthwise separable convolution to extract spatial features from the normalized IMU measurements. The LKA is applied to learn long-range relationships further. Numerical experiments and variation studies results prove the effectiveness of the proposed model.  






\section*{ACKNOWLEDGMENT}
The authors are very grateful to Martin Brossard for the open-source code and the results of attitude estimation.

\bibliographystyle{ieeetr}
\bibliography{ref}
\end{document}